\RequirePackage{amsmath}
\documentclass[runningheads]{llncs}
\usepackage[T1]{fontenc}
\usepackage{amssymb}
\usepackage{graphicx}
\usepackage{multirow}
\usepackage{listings}
\usepackage{makecell}
\usepackage[breaklinks=true, colorlinks, linkcolor=black, citecolor=black, urlcolor=blue]{hyperref}
\usepackage{color}
\usepackage{cellspace}
\usepackage{cleveref}

\newif\ifshowcomments
\showcommentstrue
\ifshowcomments
\newcommand{\mynote}[2]{\fbox{\bfseries\sffamily\scriptsize{#1}}
{\small$\blacktriangleright$\textsf{#2}$\blacktriangleleft$}}
\else
\newcommand{\mynote}[2]{}
\fi

\begin{document}
%
\title{Towards a Benchmark for Causal Business Process Reasoning with LLMs\texorpdfstring{\thanks{This project has received funding from the European Union's Horizon research and innovation programme under grant agreements no 101094905 (AI4GOV), 101092021 (AutoTwin), and 101092639 (FAME).}}{}}
\titlerunning{A BP$^C$ Benchmark for LLMs}
%
\author{Fabiana Fournier \and
Lior Limonad \and
Inna Skarbovsky}
\authorrunning{F. Fournier et al.}
%
\institute{IBM Research, Haifa, Israel\\ 
\email{\{fabiana,liorli,inna\}@il.ibm.com}}
\maketitle              
\begin{abstract}

Large Language Models (LLMs) are increasingly used for boosting organizational efficiency and automating tasks.
While not originally designed for complex cognitive processes, recent efforts have further extended to employ LLMs in activities such as reasoning, planning, and decision-making. In business processes, such abilities could be invaluable for leveraging on the massive corpora LLMs have been trained on for gaining deep understanding of such processes. In this work, we plant the seeds for the development of a benchmark to assess the ability of LLMs to reason about causal and process perspectives of business operations. We refer to this view as Causally-augmented Business Processes (BP$^C$). 
The core of the benchmark comprises a set of BP$^C$ related situations, a set of questions about these situations, and a set of deductive rules employed to systematically resolve the ground truth answers to these questions. Also with the power of LLMs, the seed is then instantiated into a larger-scale set of domain-specific situations and questions.
Reasoning on BP$^C$ is of crucial importance for process interventions and process improvement. Our benchmark, accessible at \url{https://huggingface.co/datasets/ibm/BPC}, can be used in one of two possible modalities: testing the performance of any target LLM and training an LLM to advance its capability to reason about BP$^C$.  

\end{abstract}
\section{Introduction}

Large Language Models (LLM) refers to statistical models of natural language that, based on the large corpora of text data they have been trained on, predict next plausible tokens (basic units of text) given an input string~\cite{Zhao2023}. LLMs have been successfully applied to a wide range of applications including: Chatbots and virtual assistants (e.g., automated customer support); content generation and automation (e.g., articles and blogs generation); language translation; text summarization and document analysis; and question answering.
The interaction with LLMs is conventionally attained via a textual prompt in which the content and the instructions to the LLM are being constructed, also known as \textit{prompt engineering}. 
The process is iterative, with the model’s output being analyzed and the prompt adjusted accordingly. Prompt engineering is key to the efficient use of LLMs~\cite{Teubner2023}. 
However, LLM models can only answer prompts accurately if they have been fed the right training data as they lack planning and reasoning capabilities (e.g.,~\cite{Moritz2022,Subbarao2024}). As stated by Yann LeCun\footnote{\url{https://www.ft.com/content/23fab126-f1d3-4add-a457-207a25730ad9}}, chief AI scientist of Facebook and Instagram, LLMs have ``very limited understanding of logic...do not understand the physical world, do not have persistent memory, cannot reason in any reasonable definition of the term...''. In fact, ChatGPT has a causal hallucination issue when tackling causal relationships which cannot be overcome relying solely on prompts~\cite{Gao2023}.



While reasoning may not be inherent in the architecture of LLMs, critics do argue that the ability to reason does manifest itself as an emergent property in LLMs~\cite{Wei2022}. That is, once trained on sufficiently large corpora of examples, LLMs may become capable of statistically deriving statements that are also coincidentally sound with the given set of arguments in the input. Similar emergent abilities may also include planning, decision-making, in-context learning, and answering in zero-shot settings~\cite{Humza2023}. 
Whether reasoning is an inherent capability or an emergent property of future LLMs, particularly for interpreting business process models, is likely to remain a heated debate. This debate emphasizes the importance of establishing benchmarks for evaluating LLM performance. 
Our focus is on driving such enablement by using LLMs to analyze textual descriptions of Causally-augmented Business Processes (BP$^C$s). 
Such descriptions differ from conventional process descriptions by also including statements about the causal execution dependencies among the activities as first-class citizens.
Teaching LLMs to reason about BP$^C$s could enhance the analysis and improvement of such processes.

To standardize and measure the ability of LLMs to facilitate such tasks, a designated benchmark is being developed. The purpose of the developed novel benchmark is twofold: as a testing dataset that can be employed to quantify the ability of an LLM to reason about BP$^C$s, and as a training dataset that can be used to adapt the ability of an LLM for this task. 
We evaluated two open-source and three commercial LLMs on a small subset of the situations and questions. Our results highlight the importance of producing an objective numeric scale partitioned by different perspectives to compare and assess various LLM performances. They also indicate that there is room for further improvement of LLMs in the task of reasoning about BP$^C$s. The evolving benchmark and corresponding prompts are available here: \url{https://github.com/IBM/SAX/tree/main/NLP4BPM2024}.

Given these observations, we are cautious in claiming that LLMs possess a genuine inherent ability to reason. We approach this by acknowledging that while LLMs may not inherently reason, they can achieve predictive accuracy through training on a large set of deductive textual statements. This accuracy results in syntactic output that reliably aligns with what would be produced by genuine logical reasoning. Whether reaching such a level of predictive accuracy is philosophically equivalent to actually having the ability to reason is something that we leave beyond the scope of our study. 



\vspace{-0.5em}
\section{Background}
Business process management is the discipline that combines approaches for the
design, execution, control, measurement, and optimization of business processes~\cite{vanderAalst2016ProcessMining}.
With the penetration of AI applications into organizations, the concept of AI-Augmented Business Process Management Systems (ABPMSs) was coined in~\cite{Dumas2023}. ABPMSs are process-aware information systems that rely on trustworthy AI technology to continuously adapt and improve a set of business processes with respect to one or more performance indicators. In such systems, subsymbolic AI methods are not used to replace human or symbolic reasoning in crucial tasks, but rather to support human and machine decisions and actions~\cite{Kampik2023}. A natural way of driving this interaction between humans and AI is through LLMs. To enable this, it is important for LLMs to support reasoning about business processes. In analogy to ~\cite{Huang2023}, by reasoning we refer here to the process of thinking about business processes in a logical and systematic way, using evidence and past experiences to reach a conclusion or make a decision, for the sake of process improvement. As shown in~\cite{Fournier2023TheDependencies}, understanding the temporal dependencies among the tasks in the process is not enough for reasoning about the consequences of interventions underlying process improvement decisions. To this aim, we specify here the Causally-augmented Business Process (BP$^C$) formalism as the concrete flavor for process descriptions in which not only the temporal flow but also the causal relations among the tasks are inherently captured. 

A BP$^C$ is a business process extended with inter-activity relations that reflect causal execution dependencies. A \texttt{causal execution dependence}, denoted as $A \xrightarrow{c} B$, implies that the time task $B$ executes is determined by the time task $A$ executes in a given process as defined in~\cite{Fournier2023TheDependencies}. From a process perspective, we consider~\texttt{followed-by} as a general form of a temporal relation among activities, hereafter denoted as $A \rightarrow B$, implying that according to most process observations (frequently evidenced in many process execution logs), the execution of activity $B$ occurs either immediately after the execution of $A$ or sometime after. Considering the causal process perspective, a BP$^C$ may be represented in the basic form of a graph, with nodes designating activities and edges as causal execution dependencies among these activities. Three fundamental patterns, or ``junctions''~\cite{Pearl2018Why}, can be composed to characterize any causal network: $A \xleftarrow{c} B \xrightarrow{c} C$ (confounder), $A \xrightarrow{c} B \xleftarrow{c} C$ (collider), and $A \xrightarrow{c} B \xrightarrow{c} C$ (mediator).

The potential of leveraging LLMs in the BPM field has been recently researched, for example, by analyzing which opportunities and challenges LLMs pose for the individual stages of the BPM lifecycle~\cite{vidgof2023large}, in~\cite{Busch2023}, where prompt engineering techniques are discussed as an alternative to fine-tuning a specific LLM, and in~\cite{Kampik2023} where the authors introduce the notion of a Large Process Model (LPM), an envisioned neuro-symbolic software system that integrates process management knowledge in organizations with LLMs and statistical and inference methods for the automated inference of insights and actions. 
In addition, LLMs have been researched in a wide range of tasks related to BPM, including process mining~\cite{Grohs2024,Berti2024}, automation of portions of complex tasks~\footnote{\href{https://www.infoworld.com/article/3714621/how-llms-can-help-streamline-business-processes.html}{https://www.infoworld.com/article/3714621/how-llms-can-help-streamline-business-processes.html}}, conversational process modelling~\cite{Klievtsova2023}, and explainability of business process outcomes~\cite{Fahland2024}.
To date, neither there is work on leveraging LLMs for causal reasoning about business processes, nor a relevant benchmark for testing LLMs. As an early member of this family of solutions, the PET dataset~\cite{Bellan2023} presents a benchmark for question answering, featuring an initial corpus of business process descriptions annotated with activities, gateways, actors, and flow information. 
The dataset contains 45 documents with narrative descriptions of business processes and their annotations. However, the PET descriptions do not include causal execution dependencies among the activities.
In~\cite{Ning2020} the TORQUE benchmark for temporal ordering has been investigated. The dataset encompasses 21k user-generated and fully answered temporal relation questions.
Concerning both benchmarks, ours differs in tackling the task of causal reasoning about business processes.


\vspace{-1em}
\section{Approach}

\begin{figure}
    \centering
    \includegraphics[width=1\linewidth]{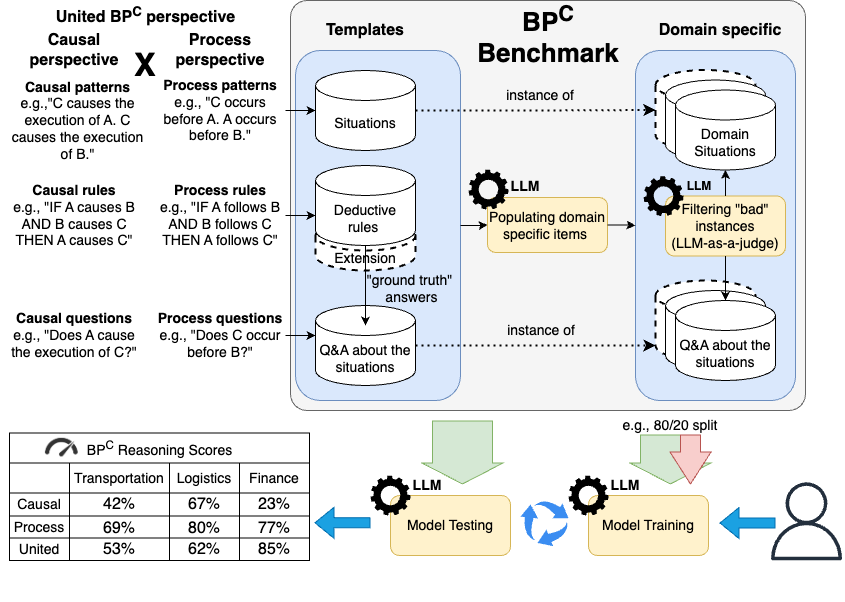}
    \vspace{-3.7em}
    \caption{BP$^C$ benchmark for testing and training of LLMs}
    \label{fig:CBP-benchmark}
    \vspace{-2em}
\end{figure}

Our goal is to develop a question-and-answer benchmark dataset to test the capacity of an LLM to produce sound answers to questions about textual narratives describing BP$^C$s.
Regardless of the debate about LLMs' suitability for reasoning tasks, we assume that with ``sufficient'' training and exposure to massive amounts of data, these models can become proficient. Yet, determining sufficiency requires some objective performance measurement.
Thus, a benchmark is necessary. We believe the developed instrumentation and methodology, shown in Figure~\ref{fig:CBP-benchmark}, are suitable for model testing and can also serve as a dataset to train LLMs to reason about BP$^C$.
The benchmark may also be further extended to more concrete domains and particular aspects of reasoning as the example given in this paper. We do acknowledge that the realization presented here is the first step in our longer journey. Hence, in addition to presenting our results so far, we also elaborate on our future evaluation intentions once the benchmark is instantiated at a greater scale.

Our approach is to define a core set of template questions and situations for basic reasoning about BP$^C$ textual narratives. These descriptions combine statements about process activities, time precedence relations, and causal execution dependencies.
At the root of such descriptions, we anticipate statements of the form ``activity $A$ occurs before activity $B$'', and ``the execution of $A$ causes the execution of $B$''. 
From a process perspective, textual descriptions typically include manifestations of the basic temporal relation \texttt{followed-by}. While there is a formal distinction between \texttt{directly-follows} and \texttt{eventually-follows}~\cite{vanderAalst2016ProcessMining}, natural language often does not strictly differentiate between them.
Hence, our work assumes a variety of manifestations of the core phrase `occurs before/after' to be expressed in the populated text. 
Similarly, from a causal process perspective, the descriptions in the text are likely to include manifestations of the causal relation of \texttt{causal-execution-dependence}. Respectively, manifestations of the core phrase of ``causes the execution of'' are populated in the text.

Therefore, to account for complete coverage of textual narratives that are descriptive of any structural form of a BP$^C$, we partition the set of template situations into three subsets corresponding to each of the three fundamental causal patterns. 
Respectively, from a process perspective, each pattern is associated with either a congruent process structure or a simple execution sequence of the form $A \rightarrow B \rightarrow C$ (as shown in~\cite{Fournier2023TheDependencies}).
The core set of situations that can be composed to describe any real-world BP$^C$ situation is shown in Table~\ref{tab:CBP-domain}. 
A situation refers to the unified temporal and causal relations among any subset of three activities $A$, $B$, and $C$. These conditions provide the core content for generating domain-specific text based on the respective template phrases listed.

\begin{table}[ht]
\centering
\caption{The domain of five different BP$^C$ situations as spanned by the three causal patterns and corresponding process structures}
\vspace{-0.6em}
\label{tab:CBP-domain}
\resizebox{\textwidth}{!}{%
\begin{tabular}{|Sl|cc|cc|c|}
\hline
 & \multicolumn{2}{Sc|}{Confounder} & \multicolumn{2}{c|}{Collider} & \multicolumn{1}{c|}{Mediator} \\ \hline
Situation$\#$ & \multicolumn{1}{c|}{1} & \multicolumn{1}{c|}{2} & \multicolumn{1}{c|}{3} & \multicolumn{1}{c|}{4} & \multicolumn{1}{c|}{5} \\ \hline
\begin{tabular}[c]{@{}l@{}}Causal\\ structure\end{tabular} & \multicolumn{2}{c|}{$A\xleftarrow{C}B\xrightarrow{C}C$} & \multicolumn{2}{c|}{$A\xrightarrow{C}B\xleftarrow{C}C$} & \multicolumn{1}{c|}{$A\xrightarrow{C}B\xrightarrow{C}C$} \\ \hline
\begin{tabular}[c]{@{}l@{}}Process\\ structure\end{tabular} & \multicolumn{1}{l|}{$A\leftarrow B\rightarrow C$ (split)} & \begin{tabular}[c]{@{}l@{}}$B\rightarrow A\rightarrow C$\\ (or $B\rightarrow C\rightarrow A$)\end{tabular} & \multicolumn{1}{l|}{$A\rightarrow B\leftarrow C$ (join)} & \begin{tabular}[c]{@{}l@{}}$A\rightarrow C\rightarrow B$\\ (or $C\rightarrow A\rightarrow B$)\end{tabular} & $A\rightarrow B\rightarrow C$ \\ \hline
\begin{tabular}[c]{@{}l@{}}Causal phrase\\ template\end{tabular} & \multicolumn{2}{c|}{\begin{tabular}[c]{@{}l@{}}B causes the execution of A,\\ B causes the execution of C\end{tabular}} & \multicolumn{2}{c|}{\begin{tabular}[c]{@{}l@{}}A causes the execution of B,\\ C causes the execution of B\end{tabular}} & \begin{tabular}[c]{@{}l@{}}A causes the execution of B,\\ B causes the execution of C\end{tabular} \\ \hline
\begin{tabular}[c]{@{}l@{}}Process phrase\\ template\end{tabular} & \multicolumn{1}{l|}{\begin{tabular}[c]{@{}l@{}}B occurs before A,\\ B occurs before C\end{tabular}} & \begin{tabular}[c]{@{}l@{}}B occurs before A,\\ A occurs before C\end{tabular} & \multicolumn{1}{l|}{\begin{tabular}[c]{@{}l@{}}A occurs before B,\\ C occurs before B\end{tabular}} & \begin{tabular}[c]{@{}l@{}}A occurs before C,\\ C occurs before B\end{tabular} & \begin{tabular}[c]{@{}l@{}}A occurs before B,\\ B occurs before C\end{tabular} \\ \hline
\end{tabular}%
}
\vspace{-2em}
\end{table}

Respectively, we associate each situation with a set of deductive rules to facilitate the reasoning about each situation. The rules are split between causal, process, and the combination of the two perspectives, capturing the meaning of the relations in each situation, unfolding from fundamental properties of the relations in discrete mathematics, jointly with a closed world assumption according to which any unknown premise is deduced to be \texttt{False}. 
For each situation, we generated a set of ``Yes''/``No'' template questions answerable by deductive reasoning that combines the facts stated in the situation with at least one associated rule. 
For example, the basic rule of transitivity on the relation of causal execution dependence, i.e., $A \xrightarrow{c} C \Leftarrow (A \xrightarrow{c} B) \land (B \xrightarrow{c} C)$, may be employed to resolve the answer to the question ``Does A cause the execution of C?'' when applied in the context of Mediator situation \texttt{\#}5 (Table~\ref{tab:CBP-domain}). 

For brevity, we include here as an example the set of rules (see Table~\ref{tab:confounder-rules}) and corresponding template questions (see Table~\ref{tab:confounder-questions}) created for situation \texttt{\#}2. This situation addresses a BP$^C$ condition in which the confounder pattern in the causal perspective is associated with a sequence pattern of the same three activities in the process perspective. Concerning symmetry and reflexivity rules (Table~\ref{tab:confounder-rules}), 
it is assumed that any cycle in the BP$^C$ structure can be entangled with a method such as k-loop unrolling~\cite{Tanmayee2019}.

\begin{table}[ht]
\vspace{-1.5em}
\centering
\caption{Deductive rules associated with the confounder situation \texttt{\#}2, considering the manifestation of $\rightarrow$ as ``occurs before'' and of $\xrightarrow{c}$ as ``causes the execution of''.}
\vspace{-0.6em}
\label{tab:confounder-rules}
\resizebox{\textwidth}{!}{%
\begin{tabular}{|Sl|Sl|Sl|}
\hline
 & Rules & Comments \\ \hline
\multirow{4}{*}{Process related} & PR1: $A\rightarrow C \Leftarrow (A\rightarrow B) \land (B\rightarrow C)$ & Transitivity of the $\rightarrow$ relation. \\ \cline{2-3} 
 & PR2: $B\not\rightarrow A \Leftarrow A\rightarrow B$ & Asymmetry of the $\rightarrow$ relation. \\ \cline{2-3} 
 & PR3: $A\leftarrow B \Leftarrow B\rightarrow A$ & \begin{tabular}[c]{@{}l@{}}Where $\leftarrow$ is an antonym manifestation of $\rightarrow$,\\ e.g., ``occurs after''.\end{tabular} \\ \cline{2-3}
 & PR4: $A \not\rightarrow A$ & No reflexivity. \\ \cline{2-3}
 & PR5: $ B\Leftarrow (A\rightarrow B) \land A$ & \begin{tabular}[c]{@{}l@{}}Entailed from the meaning of $\rightarrow$:\\ i.e., if $A$ executes then $B$ will execute at some later time.\end{tabular} \\ \hline
\multirow{4}{*}{Causal related} & CR1: $B\not\xrightarrow{c} A \Leftarrow A\xrightarrow{c} B$ & Asymmetry of the $\xrightarrow{c}$ relation. \\ \cline{2-3}
 & CR2: $A\xrightarrow{c} C \Leftarrow (A\xrightarrow{c} B) \land (B\xrightarrow{c} C)$ & \begin{tabular}[c]{@{}l@{}}Transitivity of the $\xrightarrow{c}$ relation.\\ Relevant for situation $\#$5.\end{tabular} \\ \cline{2-3} 
 & CR3: $A\xleftarrow{c} B \Leftarrow B\xrightarrow{c} A$ & \begin{tabular}[c]{@{}l@{}}Where $\xleftarrow{c}$ is an antonym manifestation of $\xrightarrow{c}$,\\ e.g., ``because the execution of''.\end{tabular} \\ \cline{2-3} 
 & \begin{tabular}[c]{@{}l@{}} CR4: $\lnot B\Leftarrow (A\xrightarrow{c} B) \land \lnot A$ and\\ $ A\Leftarrow (A\xrightarrow{c} B) \land B$ \end{tabular}  & \begin{tabular}[c]{@{}l@{}}Entailed from the meaning of $\xrightarrow{c}$:\\ i.e., if $A$ doesn't execute then $B$ doesn't execute (and vice versa).\end{tabular} \\ \hline
\multirow{2}{*}{\begin{tabular}[c]{@{}l@{}}Process and\\ causal structure\end{tabular}} & PCR1: $A \rightarrow B \Leftarrow A\xrightarrow{c} B$ & Causal execution dependence implies time precedence in the process. \\ \cline{2-3} 
 & PCR2: $B \not\xrightarrow{c} A \Leftarrow A\rightarrow B$ & \begin{tabular}[c]{@{}l@{}}Time precedence in the process implies no causal execution\\ dependence on the opposite direction.\end{tabular} \\ \hline
\end{tabular}%
}
\vspace{-2em}
\end{table}

Table~\ref{tab:confounder-questions} shows for each of the populated template questions, which of the rules in Table~\ref{tab:confounder-rules} were involved in deducing its answer w.r.t the condition articulated in situation \texttt{\#}2. These answers to the template questions remain as the ``ground truth'' for their corresponding instantiated domain-specific versions. 

\begin{table}[htb]
\centering
\caption{A core set of template questions with answers deduced by the rules for the confounder situation \texttt{\#}2}
\vspace{-0.6em}
\label{tab:confounder-questions}
\resizebox{\textwidth}{!}{%
\begin{tabular}{|l|l|l|l|}
\hline
 & Template question & Answer & Related rule(s) \\ \hline
\multirow{7}{*}{\begin{tabular}[c]{@{}l@{}}Process \\ related\end{tabular}} & QP1: Does C occur before B? & Yes & PR1 \\ \cline{2-4} 
 & QP2: Does B occur before C? & No & PR2 and QP1 \\ \cline{2-4} 
 & QP3: Does A occur after C? & Yes & PR3 \\ \cline{2-4} 
 & QP4: Does B occur after C? & Yes & PR3 and QP1 \\ \cline{2-4} 
 & QP5: Does C occur after A? & No & PR2 \\ \cline{2-4} 
 & QP6: Does C occur after B? & No & PR2 and QP1 \\ \cline{2-4} 
 & QP7: Does A occur after B? & No & PR3 and PR2 \\ \hline
\multirow{6}{*}{\begin{tabular}[c]{@{}l@{}}Causal \\ related\end{tabular}} & QC1: Does A cause the execution of C? & No & CR1 \\ \cline{2-4} 
 & QC2: Does B cause the execution of C? & No & CR1 \\ \cline{2-4} 
 & QC3: Does A execute because of C? & Yes & CR3 \\ \cline{2-4} 
 & QC4: Does B execute because of C? & Yes & CR3 \\ \cline{2-4} 
 & QC5: If C doesn’t execute, will A ever execute? & No & CR4 \\ \cline{2-4} 
 & QC6: If C doesn’t execute, will B ever execute? & No & CR4 \\ \hline
\multirow{2}{*}{\begin{tabular}[c]{@{}l@{}}Process and \\ causal related\end{tabular}} & QPC1: Does A cause the execution of B? & No & \begin{tabular}[c]{@{}l@{}}Close world\\ assumption\end{tabular} \\ \cline{2-4} 
 & QPC2: Does B cause the execution of A? & No & PCR2 \\ \hline
\end{tabular}%
}
\vspace{-3em}
\end{table}

Similar to the underlying set of situations, domain-specific questions were methodologically instantiated from these questions. While the core set of template questions is extensible (see section~\ref{sec:extension}), any specific set uniquely tags the benchmark version 
and also sets a bound to its expressiveness, that is, the grammatical richness the benchmark can accommodate. 
Pragmatically, this means that any two target LLMs can be compared only when assessed according to the same benchmark version.
\vspace{-1.2em}
\subsection{Extending the benchmark}
\label{sec:extension}

As noted, the core set of questions may be extended with additional template questions 
to assess the capacity of an LLM to reason about any additional aspect of interest. 
For example, we may define the boolean function \texttt{is\_shortened}($A$) to denote that the execution time of activity $A$ was expedited to finish its execution earlier. Respectively, the following rule can be added: 
\setlength{\abovedisplayskip}{3pt}
\setlength{\belowdisplayskip}{3pt}
\[CR5: is\_shortened(B) \Leftarrow A\xrightarrow{c}B \land is\_shortened(A) \]
Similarly, other functions can be added to denote other forms of temporal intervention, such as \texttt{is\_extended}($A$), \texttt{is\_halted}($A$), and \texttt{is\_delayed}($A$). 

Respective to adding this rule, the causal set of questions for situation \texttt{\#}2 above can be extended with the additional template questions as listed in Table~\ref{tab:confounder-questions-extension}.

Another form of extension can be achieved by extending the grammar underlying the process situations to express more complex structures (e.g., gateways).

\begin{table}[htb]
\vspace{-1em}
\centering
\caption{An extension to the core set of template questions for situation \texttt{\#}2.}
\vspace{-0.6em}
\label{tab:confounder-questions-extension}
\resizebox{\textwidth}{!}{%
\begin{tabular}{|l|l|l|l|}
\hline
 & Template question & Answer & Related rule(s) \\ \hline
\multirow{4}{*}{\begin{tabular}[c]{@{}l@{}}Causal \\ related\end{tabular}} & QC7: If we shorten A, will B be shortened? & No & CR5 \\ \cline{2-4} 
 & QC8: If we shorten B, will A be shortened? & No & CR5 \\ \cline{2-4} 
 & QC9: If we shorten C, will A be shortened? & Yes & CR5 \\ \cline{2-4} 
 & QC10: If we shorten C, will B be shortened? & Yes & CR5 \\ \hline
\end{tabular}%
}
\vspace{-3em}
\end{table}

\subsection{Populating domain-specific questions}
\vspace{-1em}
We used the open-source Mixtral-instruct-8x-7b\footnote{\url{https://huggingface.co/mistralai/Mixtral-8x7B-Instruct-v0.1\#model-card-for-mixtral-8x7b}\label{fn:mixtral}} LLM to instantiate the template questions for situation~\texttt{\#}2 to concrete problem domain statements for each of the three perspectives. For this work, we restricted the instantiation of the 15 questions in Table~\ref{tab:confounder-questions} and the additional 4 in Table~\ref{tab:confounder-questions-extension}, to having each template question populated with one corresponding domain-specific question. 
The domains were arbitrarily selected as one of the following: transportation, manufacturing, logistics, retail, finance, insurance, and medical. An example of a prompt employed for such instantiation attending to the process perspective in situation~\texttt{\#}2 (QP1) is illustrated as prompt~\texttt{\#}1 in Table~\ref{tbl:llm-prompts}.

\begin{table}[htb]
\vspace{-0.8em}
\centering
\caption{Series of LLM prompts used during benchmark development}
\vspace{-0.8em}
\label{tbl:llm-prompts}
\begin{tabular}{|p{0.015\linewidth}|p{0.985\linewidth}|}
\hline
\texttt{\#} & \textbf{LLM prompt} \\
\hline
1 & \makecell[l]{\textbf{INPUT:}\\ \textbf{Phrase:} A, B, and C are activities in some process.\\ 
C occurs before A. A occurs before B.\\
\textbf{Question:} Does C occur before B?\\
\textbf{Instruction:} Considering the above question, choose any relevant concrete\\ activities A, B, and C in the domain of $\ll$transportation$\gg$ that retain the truth of\\ the phrase statements. Using these terms, instantiate corresponding phrase\\ statements and a question statement in a form that matches the above statements\\ and question.\\
\textbf{OUTPUT:}\\
\textbf{Concrete Activities:}\\
A: Boarding the plane B: Takeoff of the plane C: Check-in at the airport\\
\textbf{Instantiated Phrase Statements:} Check-in at the airport occurs before boarding\\ the plane. Boarding the plane occurs before takeoff of the plane.\\
\textbf{Instantiated Question Statement:} Does check-in at the airport occur before\\ takeoff of the plane?
} \\
\hline
2 & \makecell[l]{\textbf{INPUT:}\\
\textbf{Template question:} If we shorten C, will B be shortened?\\ 
\textbf{Instantiated question:} If we shorten \textit{Damage assessment team inspects the}\\ \textit{damage and estimates the cost of repair} in duration, will \textit{Insurance policyholder}\\ \textit{receives the payout} be shortened?\\
\textbf{Instruction:} Considering the instantiated question above a concrete version of the\\ template question where the letters are replaced with process activity descriptions,\\ when replacing these descriptions with their corresponding letters in the template,\\ how would you rate the similarity between the revised instantiated question and the\\ template question (where a 1 rate means they are identical and 0 they are\\ completely different)? In your output, print only the rate value on a 0-1 scale.\\
\textbf{OUTPUT:}\\
0.9 or 90\% similarity.} \\
\hline
\end{tabular}
\vspace{-2.7em}
\end{table}

\subsection{Filtering out inadequate questions}

Our instantiated set of questions so far was relatively small. 
However, in full-scale development, covering the complete situation space and more domains, the number of questions will likely increase significantly.
This process might generate domain-specific questions in a form that becomes incongruent with the original template questions used as its seed due to the non-deterministic nature of LLMs. Therefore, we also foresee the use of ``LLM-as-a-judge'' to curate the quality of the generated questions. That is, excluding the ones that do not maintain a faithful linkage to their corresponding template. To this end, prompt\texttt{\#}2 in Table~\ref{tbl:llm-prompts} 
was created to grade and remove questions scoring below an acceptable threshold.

\vspace{-0.8em}
\section{Evaluation}


An initial instantiation of the benchmark seed, the `benchmark prototype' dataset, was populated for testing purposes, including one domain-specific question for each template question in situation \#2. We used this prototype to assess its applicability and measure accuracy against state-of-the-art LLMs across three perspectives: process, causal, and their combination.

At first, we ran the three core sets of template questions from Table~\ref{tab:confounder-questions} with five different target LLMs: GPT3.5\footnote{\url{https://beta.openai.com/docs/models/gpt-3}}, GPT4\footnote{\url{https://www.openai.com/research/gpt-4}}, GPT4o\footnote{\url{https://openai.com/index/hello-gpt-4o/}}, Mixtral-instruct-8x-7b$^{\ref{fn:mixtral}}$, and Merlinite-7b\footnote{\url{https://huggingface.co/ibm/merlinite-7b}}. We repeated each template question ten times with clean-slate prompts, preceding each question with the corresponding phrase describing the situation and instruction as illustrated in Table~\ref{tbl:llm-prompts-tests}. We averaged the proportion of correctly answered questions for each reasoning perspective: process, causal, and combined. We then also added the extended set of questions (see section~\ref{sec:extension}) and revised the results accordingly.

As a second step, using the domain-specific questions, we repeated the benchmark testing for the two open-source LLMs (Mixtral-instruct-8x-7b and Merlinite-7b) ten times per prompt, measuring the proportion of correct answers for the three perspectives, both without and with the inclusion of the extension questions.

\begin{table}[htb]
\centering
\caption{LLM Prompts used for benchmark testing}
\vspace{-1.1em}
\label{tbl:llm-prompts-tests}
\begin{tabular}{|p{0.015\linewidth}|p{0.985\linewidth}|}
\hline
\texttt{\#} & \textbf{LLM prompts: process, causal, and both} \\
\hline
1 & \makecell[l]{\textbf{Phrase:} A, B, and C are activities in some process. \\ $\ll$process related$\gg$C occurs before A. A occurs before B.\\
$\ll$causal related$\gg$C causes the execution of A. C causes the execution of B.\\
\textbf{Instruction:} Considering the above phrase about activities in a process, answer\\ the following question. Your answer should be limited to either Yes or No and\\ nothing else.\\
\textbf{Question:}\\ 
$\ll$process related (QP1)$\gg$Does C occur before B?\\
$\ll$causal related (QC1)$\gg$Does A cause the execution of C?\\
$\ll$causal \& process related (QPC1)$\gg$Does A cause the execution of B?
} \\
\hline
\end{tabular}
\vspace{-2.5em}
\end{table}


We acknowledge that our current evaluation caters strictly to the applicability of the benchmark to a handful of LLMs. Such an assessment lacks characterizing the benchmark quality. For this, we elaborate here on a set of relevant metrics that we intend to assess once the benchmark is developed in full scale.

\textbf{Completeness} refers to the range of realistic BP$^C$ situations the benchmark covers. This is ensured by using causal ``junctions'' as per~\cite{Pearl2018Why}, which accommodate any causal structure, and by the richness of rules and their coverage by questions. Our design principle ensures each rule helps resolve at least one question and that the benchmark is extensible. A metric can capture the proportion of rules covered by questions. However, completeness is always limited by the core set of template questions and the finite domains involved. Therefore, it is crucial to disclose the list of questions and domains for any benchmark version.

\textbf{Correctness} refers to the degree each question's answer is adequate to the targeted process situation. For Yes/No questions, a correct answer indicates whether the condition holds (or not) in the corresponding situation. We use rules as a formal mechanism to determine answers and keep the same answers for the instantiated situations and corresponding questions.
    

\textbf{Reliability} reflects result consistency across multiple tests. This may also be influenced by the LLM's inherent consistency. It can be measured using conventional metrics like Cronbach's alpha. To ensure reliability, we reset the prompt to prevent prior context from affecting interactions.

\textbf{Validity} assesses how well each question captures the specific aspect of its domain. This can be measured through convergence metrics like factor loading, and discriminant validity when partitioning the instantiated questions by different perspectives and domains.

\vspace{-0.8em}
\section{Results}
\vspace{-0.5em}

We report the results for running our benchmark prototype in Table~\ref{tab:accuracy-benchmark-results}. This table shows 
the proportion of questions answered correctly by each LLM. 
The results are split between the template questions 
and domain-specific questions, and are also partitioned by the various perspectives, considering the causal perspective both without and with the addition of the extending questions. 
While the template questions provide exhaustive domain coverage, the domain-specific ones should be interpreted cautiously, as each template question has only one corresponding domain-specific question.
As such, we expect that in a full-scale set of domain-specific questions, the performance is likely to get closer to the accuracy presented by the template questions. In addition, it is more likely the corpora employed for LLM training were domain-specific, hence incidentally implying an improved performance for domain-specific questions.

\begin{table}[htb]
\centering
\caption{LLM accuracy results using the prototype benchmark for situation~\texttt{\#}2}
\vspace{-0.9em}
\label{tab:accuracy-benchmark-results}
\resizebox{\textwidth}{!}{%
\begin{tabular}{lc|ccccc|cc}
 &  & \multicolumn{5}{c|}{\textbf{Template questions}} & \multicolumn{2}{c}{\textbf{Domain-specific questions}} \\ \hline
 & \textbf{\begin{tabular}[c]{@{}c@{}}Num of \\ questions\end{tabular}} & \textbf{\begin{tabular}[c]{@{}c@{}}Merlinite\\ 7b\end{tabular}} & \textbf{\begin{tabular}[c]{@{}c@{}}Mixtral \\ instruct \\ 8x 7b\end{tabular}} & \textbf{\begin{tabular}[c]{@{}c@{}}GPT\\ 4o\end{tabular}} & \textbf{\begin{tabular}[c]{@{}c@{}}GPT\\ 4\end{tabular}} & \textbf{\begin{tabular}[c]{@{}c@{}}GPT\\ 3.5\end{tabular}} & \textbf{\begin{tabular}[c]{@{}c@{}}Merlinite\\ 7b\end{tabular}} & \textbf{\begin{tabular}[c]{@{}c@{}}Mixtral \\ instruct \\ 8x 7b\end{tabular}} \\ \hline
Process & 7 & 58\% & 71\% & 100\% & 85\% & 65\% & 100\% & 100\% \\
Causal & 6 & 66\% & 100\% & 100\% & 100\% & 56\% & 98\% & 100\% \\
Process + Causal & 2 & 100\% & 100\% & 100\% & 80\% & 15\% & 55\% & 95\% \\
Extension & 4 & 50\% & 50\% & 70\% & 65\% & 55\% & 73\% & 50\% \\
Causal+Extension & 10 & 60\% & 80\% & 88\% & 86\% & 56\% & 88\% & 80\% \\ \hline
\textbf{Total weighted avg} &  & \textbf{63\%} & \textbf{79\%} & \textbf{94\%} & \textbf{85\%} & \textbf{55\%} & \textbf{89\%} & \textbf{89\%}
\end{tabular}%
}
\vspace{-1em}
\end{table}

The benchmark can be used for testing LLM performance, as demonstrated here, and for training an LLM to improve its reasoning about BP$^C$. For testing, question answers serve as the ``ground truth'' to determine model accuracy, which can be analyzed by perspective and domain. For training, the questions and answers can be randomly split into training and testing subsets (e.g., 80/20). The testing subset is used before and after training to measure improvement.

\section{Conclusion and Future Work}

Setting aside the debate whether LLMs can generally reason, our goal here is more modest. We would like to equip (or measure) LLMs regarding their capacity to adequately infer sound conclusions when presented with knowledge about causal business processes. In this regard, our developed instrumentation can be employed in two (complementary) manners. It could be used to test such an ability with respect to a relatively wide variety of process domains. 
In addition, the tool can also be used as a model training dataset to augment an existing LLM with such ability and also to be adopted to additional problem domains for specific needs. For the former purpose, it could be used ``as is'' with only the questions component to facilitate model bench-marking, and with the ground truth answers for model training. For any newly embarked problem domains, the core seed of the template questions should be instantiated methodologically in the same process reported here to derive a corresponding set of domain-relevant items. This is also the rationale underlying our aim to release the model as open source, 
letting the community gradually contribute to the dataset to make its domain coverage broader.
Our current version of the developed benchmark can be accessed here: \url{https://huggingface.co/datasets/ibm/BPC}.

The core contribution of this work is twofold. An open-source model that is developed and a methodology reporting how to construct a benchmark for a particular task to be facilitated by LLMs, in this case, the one of reasoning about BP$^C$s.
At the time of submitting this paper, we embark on completing the specification of the template rules and questions corresponding to the other situations. As next steps, we plan to populate our benchmark with a more exhaustive set of domains and at a larger scale of instances per domain. 
Our vision is that LLM benchmarks may become standardized means to guide the choice of suitability of LLMs to specific tasks.
%
%
%

\vspace{-0.9em}
\bibliographystyle{splncs04-no-url-doi}
\bibliography{references}
%
\end{document}